\documentclass[10pt,twocolumn,letterpaper]{article}

\usepackage{cvpr}      
\usepackage{bbm}

\usepackage[dvipsnames]{xcolor}

\usepackage{floatrow}

\newcommand{\CA}{unsupervised instance segmentation model\xspace}
\newcommand{\Ours}{U2Seg\xspace}

\newcommand\footnoteWithoutNumber[1]{%
  \begingroup
  \renewcommand\thefootnote{}\footnote{#1}%
  \addtocounter{footnote}{-1}%
  \endgroup
}

\newcommand{\tablestyle}[2]{\setlength{\tabcolsep}{#1}\renewcommand{\arraystretch}{#2}\centering\footnotesize}
\definecolor{Gray}{gray}{0.9}
\usepackage{xcolor,colortbl}
\usepackage{array}
\usepackage{pifont}
\usepackage{booktabs}
\usepackage{xcolor}
\usepackage{algorithm}
\usepackage{algpseudocode}
\newcommand{\cmark}{\text{\ding{51}}}
\newcommand{\xmark}{\text{\ding{55}}}
\newlength\savewidth\newcommand\shline{\noalign{\global\savewidth\arrayrulewidth
  \global\arrayrulewidth 1pt}\hline\noalign{\global\arrayrulewidth\savewidth}}
\usepackage{multirow}
\definecolor{cvprblue}{rgb}{0.21,0.49,0.74}
\usepackage[pagebackref,breaklinks,colorlinks,citecolor=cvprblue]{hyperref}

\def\paperID{241} 
\def\confName{CVPR}
\def\confYear{2024}

\title{Unsupervised Universal Image Segmentation}

\author{
\begin{tabular}{c}
Dantong Niu\footnotemark[1] \footnotemark[2] \quad
Xudong Wang\footnotemark[1] \footnotemark[2] \quad
Xinyang Han\footnotemark[1] \quad
Long Lian \quad 
Roei Herzig \quad 
Trevor Darrell
\end{tabular} \\
Berkeley AI Research, UC Berkeley\\
\small{Code:} \href{https://github.com/u2seg/U2Seg}{\small{https://github.com/u2seg/U2Seg}}
}

\begin{document}
\pagenumbering{gobble}

\twocolumn[{%
  \renewcommand\twocolumn[1][]{#1}%
  \maketitle
    \vspace{-16pt}
    \captionsetup{type=figure}
    \centering
    \includegraphics[width=0.98\textwidth]{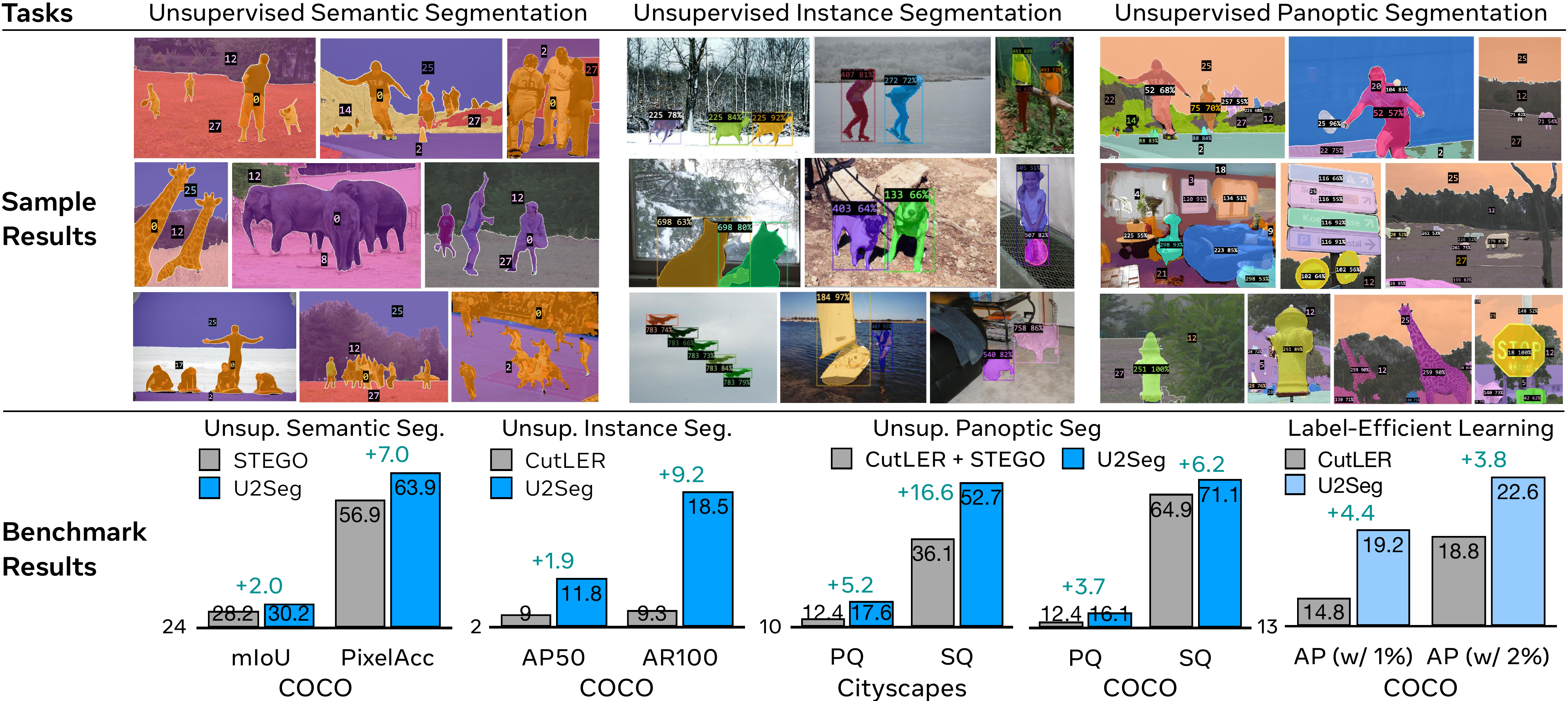}
    \vspace{-8pt}
    \caption{
    We present \textbf{\Ours}, a unified framework for \textbf{U}nsupervised \textbf{U}niversal image \textbf{Seg}mentation that consistently outperforms previous state-of-the-art methods designed for individual tasks: CutLER~\citep{wang2023cut} for unsupervised instance segmentation, STEGO~\citep{hamilton2022unsupervised} for unsupervised semantic segmentation, and the naive combination of CutLER and STEGO for unsupervised panoptic segmentation. We visualize instance segmentation results with ``semantic label'' + confidence score and semantic predictions with ``semantic label''. Zoom in for the best view. 
    }
    \label{fig:teaser}
    \vspace{14pt}
}]

\footnoteWithoutNumber{*Equal Contribution.} \footnoteWithoutNumber{$^\dagger$Project Lead.}
\begin{abstract}
Several unsupervised image segmentation approaches have been proposed which eliminate the need for dense manually-annotated segmentation masks; current models separately handle either semantic segmentation (\eg, STEGO) or class-agnostic instance segmentation (\eg, CutLER), but not both (\ie, panoptic segmentation).
We propose an Unsupervised Universal Segmentation model (U2Seg) adept at performing various image segmentation tasks---instance, semantic and panoptic---using a novel unified framework. 
U2Seg generates pseudo semantic labels for these segmentation tasks via leveraging self-supervised models followed by clustering; each cluster represents different semantic and/or instance membership of pixels. 
We then self-train the model on these pseudo semantic labels, yielding substantial performance gains over specialized methods tailored to each task: 
a {+2.6 AP$^{\text{box}}$} boost (\textit{vs.} CutLER) in unsupervised instance segmentation on COCO and a {+7.0 PixelAcc} increase (\textit{vs.} STEGO) in unsupervised semantic segmentation on COCOStuff. Moreover, our method sets up a new baseline for unsupervised panoptic segmentation, which has not been previously explored. 
U2Seg is also a strong pretrained model for few-shot segmentation, surpassing CutLER by {+5.0 AP$^{\text{mask}}$} when trained on a low-data regime, \eg, only 1\% COCO labels.
We hope our simple yet effective method can inspire more research on unsupervised universal image segmentation. 
\end{abstract}

\def\tabExtraVis#1{
\begin{figure*}[#1]
  \centering
  \includegraphics[width=0.9\linewidth]{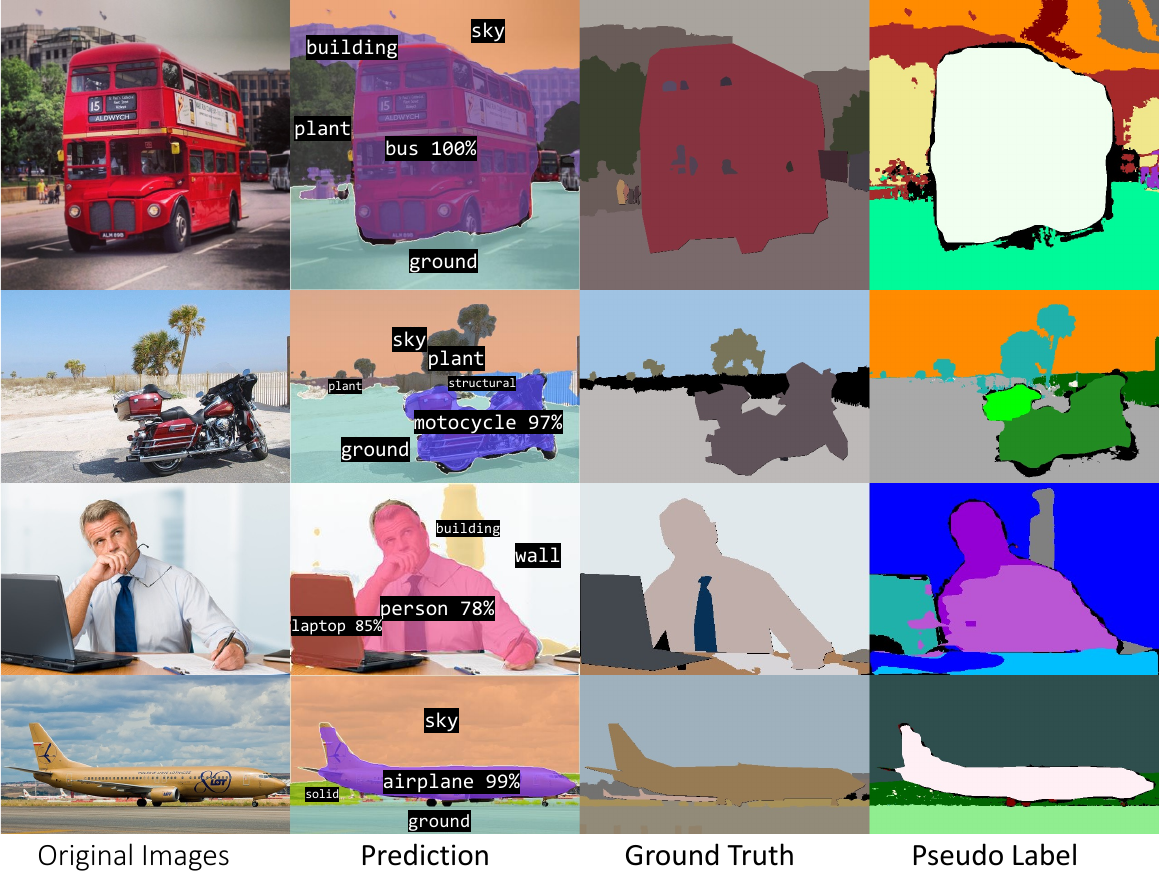}
  \caption{Universal Segmentation visualizations of \Ours's predictions.}
  \label{fig:extra-visual}
\end{figure*}
}

\section{Introduction}


The field of image segmentation has witnessed significant advancements in the recent years~\citep{he2017mask,lin2017focal,cai2018cascade,duan2019centernet,carion2020end,cheng2021maskformer,li2022exploring,kirillov2023segment}. 
Nonetheless, the effectiveness of these segmentation methods heavily depends on the availability of extensive densely human-labeled data for training  these models, which is both labor-intensive and costly and thus less scalable. 
In this paper, our objective is to explore the extent to which unsupervised image segmentation can be achieved without relying on any human-generated labels.

Several recent works such as CutLER~\citep{wang2023cut} and STEGO~\citep{hamilton2022unsupervised} have emerged as promising approaches for unsupervised image segmentation. 
CutLER leverages the property of the self-supervised model DINO~\cite{caron2021emerging} to `discover' objects without supervision, and learns a state-of-the-art localization model on pseudo instance segmentation masks produced by MaskCut~\cite{wang2023cut} (based on Normalize Cuts~\cite{shi2000normalized}).
Similarly leveraging DINO~\cite{caron2021emerging}, STEGO~\cite{hamilton2022unsupervised} introduces a novel framework that distills unsupervised features into discrete semantic labels. 
This is achieved using a contrastive loss that encourages pixel features to form compact clusters while preserving their relationships across the corpora~\cite{hamilton2022unsupervised}.
However, these methods have limitations:

\begin{itemize}[leftmargin=*,nosep]
    \item  The output of \textit{\textbf{unsupervised instance segmentation}} methods such as  CutLER~\citep{wang2023cut}  comprises \textit{class-agnostic} segments for ``things'', ignoring the ``stuff'' categories that represent pixel semantics. 
    Moreover, CutLER often treats several \textit{overlapping instances} as one instance, especially when these instances belong to the same semantic class. 
    \item On the other hand, \textit{\textbf{unsupervised semantic segmentation}} methods such as STEGO~\citep{hamilton2022unsupervised} focus on the segmentation of semantically coherent regions, lacking the capability to distinguish between individual instances.
\item \textit{\textbf{Unsupervised panoptic segmentation}} has not been addressed. Supervised panoptic segmentation methods ~\cite{Kirillov2018PanopticS,jain2023oneformer,cheng2021maskformer} predict both ``stuff'' and ``things'' classes simultaneously; to the best of our knowledge there has not been  work on unsupervised panoptic segmentation heretofore. 
\end{itemize}

To address these limitations, we propose \textbf{\Ours}, a novel \textbf{\textit{U}}nsupervised \textbf{\textit{U}}niversal image \textbf{\textit{Seg}}mentation model.
\Ours offers comprehensive scene understanding--instance, semantic and panoptic--without relying on human annotations, 
segmenting semantically meaningful regions in the image as well as 
identifying and differentiating between individual instances within those regions.

\Ours is comprised of three steps. 
First, we create high-quality, discrete semantic labels for instance masks obtained from MaskCut and DINO, by clustering semantically similar instance masks into distinct fine-grained clusters, as described in Sec.~\ref{sec:g-cutler}. 
Next, we amalgamate the semantically pseudo-labeled ``things'' pixels (from the first step) with ``stuff'' pixels (from STEGO) to produce pseudo semantic labels for each pixel in the image. 
Lastly, a universal image segmentation model is trained using these pseudo-labels, resulting in a model capable of simultaneously predicting pixel-level  (\ie, semantic segmentation and class-agnostic instance segmentation) and instance-level semantic labels, detailed in Sec.~\ref{sec:uni-cutler}.


Despite the inherent noise in these pseudo-labels, self-training the model with them yields substantial performance gains over specialized methods tailored to each task:
\Ours achieves a \textbf{+2.6 AP$^{\text{box}}$} boost (\textit{vs.} CutLER) in unsupervised instance segmentation on COCO and a \textbf{+7.0 PixelAcc} increase (\textit{vs.} STEGO) in unsupervised semantic segmentation on COCOStuff. 
Moreover, our method sets up a new baseline for unsupervised panoptic segmentation.
We also find that the multi-task learning framework and learning unsupervised segmentor with semantic labels enable our model to generate a more discriminative feature space, which makes it a superior representation for downstream supervised detection and segmentation tasks. When trained on a low-data regime, such as 1\% COCO labels, \Ours surpasses CutLER by \textbf{+5.0 AP$^{\text{mask}}$}.

\noindent \textbf{Contributions.} 
Our main contribution is the first universal unsupervised image segmentation model that can tackle unsupervised semantic-aware instance, semantic and panoptic segmentation tasks using a unified framework. 
We establish a suite of benchmarks  on unsupervised semantic-aware instance segmentation and panoptic segmentation, areas previously unexplored.
Despite using a single framework, we demonstrate that \Ours surpasses previous methods specialized for each task across all experimented benchmarks (instance, semantic, panoptic, \etc) and datasets (COCO, Cityscapes, UVO, VOC, \etc).
\section{Related Work}

\textbf{Self-supervised Representation Learning} focuses on feature learning from a large amount of unlabeled data without using human-made labels. 
\textit{Contrastive Learning-Based Methods} \citep{wu2018unsupervised, misra2020self,he2020momentum, chen2020simple} learn representation by comparing similar instances or different versions of a single instance while separating dissimilar ones. \textit{Similarity-Based Self-Supervised Learning} \citep{grill2020bootstrap, chen2021exploring} mainly reduces differences between different augmented versions of the same instance. \textit{Clustering-Based Feature Learning} \citep{xie2016unsupervised, asano2019self,zhuang2019local, caron2020unsupervised, wang2021unsupervised} finds natural data groups in the hidden space. 
\textit{Masked Autoencoders} \citep{doersch2015unsupervised,he2022masked, bao2021beit} learn by masking and then reconstructing masked parts of the image.

\noindent \textbf{Unsupervised Object Detection and Instance Segmentation.} DINO~\cite{caron2021emerging} shows that self-supervised learning (SSL) Vision Transformers (ViT)~\cite{dosovitskiy2020image} can reveal hidden semantic segmentation in images, which is not obvious in supervised counterparts~\cite{ziegler2022self,caron2021emerging}. 
Extending this, LOST~\cite{simeoni2021localizing} and TokenCut~\cite{wang2022tokencut} use DINO's patch features to identify main objects in images.
FreeSOLO~\cite{wang2022freesolo} performs unsupervised class-agnostic instance segmentation by creating coarse masks first, which are later improved through self-training. 
MaskDistill~\cite{van2022discovering} uses a self-supervised DINO to get initial masks from an affinity graph but only allows one mask per image during distillation, limiting multi-object detection. 
Meanwhile, CutLER~\cite{cutler} introduces the MaskCut method, which aims to identify multiple instances in a single image. Yet, MaskCut frequently consolidates overlapping instances into a single segment and lacks the capability to assign semantic labels to each instance. 

\noindent \textbf{Unsupervised Semantic Segmentation.}
IIC~\cite{iic} maximizes mutual information for clustering, while PiCIE~\cite{Cho2021PiCIEUS} uses invariance to photometric effects and equivariance to geometric transformations for segmentation. 
MaskContrast~\cite{maskcontrast} learns unsupervised semantic segmentation by contrasting features within saliency masks. 
STEGO~\cite{hamilton2022unsupervised} refines pretrained SSL visual features to distill correspondence information embedded within these features, thereby fostering discrete semantic clusters. 

\noindent \textbf{Universal Segmentation} has been introduced to deliver instance, semantic and panoptic segmentation tasks using a unified architecture~\cite{kirillov2017panoptic,cheng2021maskformer,cheng2021mask2former,jain2023oneformer,xiong19upsnet,kirillov2019panopticfpn,cheng2020panoptic,li2021fully,detr,wang2021max}. 
In this work, we propose \Ours to tackle this challenging task without relying on human-annotated data.

\noindent \textbf{Unsupervised Image Classification} methods mainly focus on providing a semantic label for each query image that can be mapped to ground truth classes by hungarian matching. SCAN~\cite{van2020scan} proposes a three-stage pipeline that includes representation learning, deep clustering, and self-labeling. NNM~\cite{dang2021nearest} enhances SCAN by incorporating local and global nearest neighbor matching. RUC~\cite{park2021improving} further improves SCAN using a robust loss as training objective. 
However, these approaches only provide one classification prediction per image, whereas our method provides classification per-instance for instance segmentation and per-pixel for semantic segmentation. 

\def\figPipeline#1{
    \captionsetup[sub]{font=small}
    \begin{figure*}[#1]
      \centering
      \includegraphics[width=0.9\linewidth]{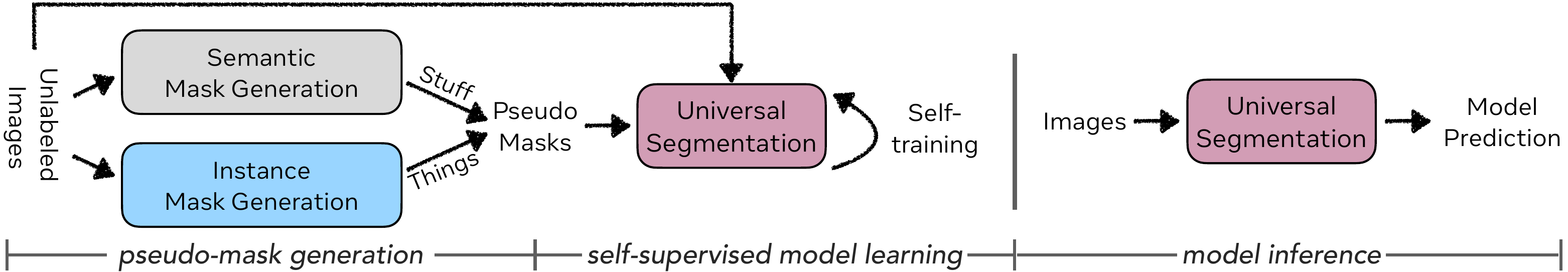}\vspace{-2pt}
      \caption{Overview of the training and inference pipeline for the proposed Unsupervised Universal Segmentation model (\Ours) adept at performing various image segmentation tasks---instance, semantic and panoptic---using a novel unified framework. 
      }
      \label{fig:pipeline}
    \end{figure*}
}

\def\figPipelineOne#1{
    \captionsetup[sub]{font=small}
    \begin{figure*}[#1]
      \vspace{-4pt}
      \centering
      \includegraphics[width=0.9\linewidth]{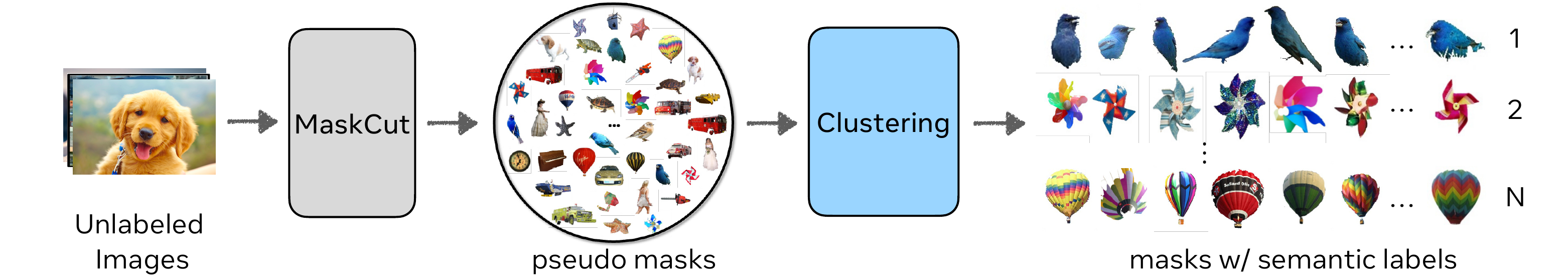}
      \vspace{-6pt}
      \caption{
      Pipeline overview for generating masks and their semantically meaningful pseudo labels in semantic-aware instance segmentation.
      We first use MaskCut to generate class-agnostic instance masks, which are then grouped into semantically meaningful clusters. 
      These pseudo semantic labels are used for training a semantic-aware instance segmentor.}
      \label{fig:pipeline-cluster-aware}
    \end{figure*}
}

\def\tabComparisonlarge#1{
\begin{table*}[#1]
\tablestyle{1.9pt}{0.9}
    \small
    \begin{center}
    \begin{tabular}{l|cc|cc|cc|cc|cc|ccc|ccc}
    
    Task $\rightarrow$      & \multicolumn{2}{c|}{Agn Instance Seg.} & \multicolumn{6}{c|}{Instance Seg.}                                             & \multicolumn{2}{c|}{Semantic Seg.} & \multicolumn{6}{c}{Panoptic Seg.}                        \\
    Datasets $\rightarrow$ & \multicolumn{2}{c|}{COCO}                  & \multicolumn{2}{c|}{COCO} & \multicolumn{2}{c|}{VOC} & \multicolumn{2}{c|}{UVO} & \multicolumn{2}{c|}{COCO}         & \multicolumn{3}{c|}{COCO} & \multicolumn{3}{c}{Cityscapes} \\
    Metric $\rightarrow$    & AP$^\text{box}$                     &     AP$_{50}^\text{box}$                & AP$_{50}^\text{box}$          & AR$_{100}^\text{box}$         & AP$_{50}^\text{box}$         & AR$_{100}^\text{box}$         & AP$_{50}^\text{box}$         & AR$_{100}^\text{box}$         & PixelAcc          & mIoU         & PQ     & SQ     & RQ     & PQ       & SQ       & RQ      \\ \shline
    FreeSOLO~\cite{wang2022freesolo}  & 9.6                 & 4.2                 & -           & -          & -          & -          & -          & -          & -                 & -            & -      & -      & -      & -        & -        & -       \\
    TokenCut~\cite{wang2022tokencut}  & 5.8                 & 3.2                 & -           & -          & -          & -          & -          & -          & -                 & -            & -      & -      & -      & -        & -        & -       \\
    CutLER~\cite{cutler}    & 21.9                & 12.3                & -           & -          & -          & -          & -          & -          & -                 & -            & -      & -      & -      & -        & -        & -       \\
    DINO~\cite{caron2021emerging}      & -                   & -                   & -           & -          & -          & -          & -          & -          & 30.5              & 9.6          & -      & -      & -      & -        & -        & -       \\
    PiCIE + H~\cite{Cho2021PiCIEUS}  & -                   & -                   & -           & -          & -          & -          & -          & -          & 48.1              & 13.8         & -       & -      & -      & -        & -        & -       \\
    STEGO~\cite{hamilton2022unsupervised}     & -                   & -                   & -           & -          & -          & -          & -          & -          & 56.9              & 28.2         & -     & -      & -      & -        & -        & -       \\ CutLER+    & -                & -                & 9.0           & 10.3          & 26.8          & 27.2          & 10.6          & 11.8          & -                 & -            & -      & -      & -      & -        & -        & -       \\
    CutLER+STEGO    & -                & -               & -           & -          & -          & -          & -          & -          & -                 & -            & 12.4      & 64.9      & 15.5      & 12.4        & 36.1       & 15.2       \\ \hline
    \rowcolor{gray!10}
    \Ours         & \textbf{22.8}               & \textbf{13.0}                & \textbf{11.8}        & \textbf{21.5}   & \textbf{31.0}       & \textbf{48.1} & \textbf{10.8}       & \textbf{25.0}       & \textbf{63.9}              & \textbf{30.2}         & \textbf{16.1}   & \textbf{71.1}   & \textbf{19.9}   & \textbf{17.6}     & \textbf{52.7}     & \textbf{21.7}    \\
    \rowcolor{gray!10}
    \textit{vs.prev.SOTA}         & \textcolor{ForestGreen}{\textbf{+0.9}}                 & \textcolor{ForestGreen}{\textbf{+0.7}}             & \textcolor{ForestGreen}{\textbf{+2.8}}           & \textcolor{ForestGreen}{\textbf{+11.2}}      & \textcolor{ForestGreen}{\textbf{+4.2}} & \textcolor{ForestGreen}{\textbf{+20.9}}   & \textcolor{ForestGreen}{\textbf{+0.2}}   & \textcolor{ForestGreen}{\textbf{+13.2}}       & \textcolor{ForestGreen}{\textbf{+7.0}}               & \textcolor{ForestGreen}{\textbf{+2.0}}          & \textcolor{ForestGreen}{\textbf{+3.7}}      & \textcolor{ForestGreen}{\textbf{+6.2}}      & \textcolor{ForestGreen}{\textbf{+4.4}}      & \textcolor{ForestGreen}{\textbf{+5.2}}        & \textcolor{ForestGreen}{\textbf{+16.6}}        & \textcolor{ForestGreen}{\textbf{+6.5}}      
    \end{tabular} \vspace{-16pt}
    \caption{With a unified framework, \Ours outperforms previous state-of-the-art methods tailored for individual tasks across various datasets, including CutLER for unsupervised instance segmentation, STEGO for unsupervised semantic segmentation, and CutLER+STEGO for unsupervised panoptic segmentation. ``Agn Instance Seg'' denotes class-agnostic instance segmentation. }
    \label{tab:comp_full}
    \end{center}
    \end{table*}
}

\figPipeline{t!}
\figPipelineOne{t!}

\section{Unsupervised Universal Segmentation}

\subsection{Preliminaries}
\label{sec:preliminaries}
We first explain the previous Unsupervised Instance Segmentation method CutLER~\cite{wang2023cut}, and Unsupervised Semantic Segmentation method STEGO~\cite{hamilton2022unsupervised}.

\noindent \textbf{CutLER}~\cite{wang2023cut} exploits self-supervised learning models like DINO~\cite{caron2021emerging} to `discover' objects and train a state-of-the-art detection and segmentation model using a cut-and-learn pipeline. 
It first uses MaskCut to extract multiple initial masks from DINO~\cite{caron2021emerging} features.
MaskCut first generates a patch-wise affinity matrix $W_{ij}\!=\!\frac{K_i K_j}{\|K_i\|_2 \|K_j\|_2}$ using the ``key'' features $K_i$ for patch $i$ from DINO's last attention layer. 
Subsequently, the cut-based clustering method Normalized Cut~\cite{shi2000normalized} is employed on the affinity matrix by finding the eigenvector $x$ that corresponds to the second smallest eigenvalue.
A foreground instance mask ${M}^{s}$ is derived through bi-partitioning of the vector $x$, enabling segmentation of individual objects in the image. For multi-instance segmentation, MaskCut iteratively refines the affinity matrix by masking out already segmented objects, allowing for subsequent extractions
\begin{align}
  W^{t}_{ij}\!=\!\frac{(K_i\prod_{s=1}^{t}{M}^s_{ij})(K_j\prod_{s=1}^{t}{M}^s_{ij})}{\|K_i\|_2\|K_j\|_2}
  \label{eqn:update-graph}
\end{align}
and repeating above steps by $N$ times. CutLER then refines detection and segmentation through a loss-dropping strategy and iterative self-training.

\noindent \textbf{STEGO}~\cite{hamilton2022unsupervised} harnesses the semantically rich feature correlations produced by unsupervised methods like DINO~\cite{caron2021emerging} for segmentation. It trains a segmentation head to refine these correlations within an image, with its K-Nearest Neighbors (KNNs), and across randomly chosen images.
Specifically, STEGO distills DINO's unsupervised features into distinct semantic labels by optimizing a correspondence loss. This loss function
measures the feature correspondences $F^{SC}$ between image feature pairs generated by DINO and the feature correspondence $S_{hwij}$ derived from a trainable, lightweight segmentation head~\cite{hamilton2022unsupervised}:
\begin{align}
L_{\text{corr}}(x, y, b) = -\sum_{hwij}(F^{SC}_{hwij} - b)\max(S_{hwij}, 0)
\label{eqn:stego-loss}
\end{align}

\subsection{Unsupervised Instance Segmentation}
\label{sec:g-cutler}

Although CutLER~\cite{wang2023cut} provides high-quality instance segmentation masks without human annotations, the predicted masks are class-agnostic, and thus do not include semantic labels for each instance. 
Our method addresses this issue by grouping the detected instances with a clustering method. 
In this way, instances assigned to the same cluster are associated with identical or closely related semantic information, while instances residing in separate clusters exhibit semantic dissimilarity.

\noindent \textbf{Pseudo Semantic Labels}. 
To train a detection and instance segmentation model, we vector quantize the model targets (pseudo semantic labels) by clustering the instance-level features of the entire dataset, under constraints derived from self-supervision. 
Specifically, our approach starts with the generation of instance segmentation masks using MaskCut~\cite{wang2023cut}. Subsequently, we utilize the efficient $K$-Means clustering method as implemented in USL~\cite{wang2022unsupervised} to cluster all segmentation masks into semantically meaningful clusters. 

We employ $K$-Means clustering to partition $n$ instances into $C (\leq n)$ clusters, where each cluster is represented by its centroid $c$~\cite{lloyd1982least, forgy1965cluster}. 
Each instance is assigned to the cluster with the nearest centroid.
Formally, we conduct a $C$-way node partitioning, denoted as $\mathcal{S} = {S_1, S_2, \ldots, S_C}$, that minimizes the within-cluster sum of squares~\cite{kriegel2017black}:
\begin{align}
\begin{split}
\min_{\mathcal{S}} \sum_{i=1}^{C} \sum_{V \in S_i} \left| V - c_i \right|^2 = \min_{\mathcal{S}} \sum_{i=1}^C |S_i| \text{Var}(S_i)
\end{split}
\label{eqn:kmeans}
\end{align}

This optimization process is carried out iteratively using the EM algorithm~\cite{mclachlan2007algorithm}, starting from selecting random samples as initial centroids. As a result, this process assigns pseudo semantic labels, denoted as $y_i$, to each instance $i$, with $y_i$ falling within the range of $[1, C]$.


The resulting semantic labels serve multiple purposes:
\textbf{\textit{1) Semantic-aware copy-paste augmentation}}, which significantly improves CutLER's capability to differentiate overlapping instances, especially when they share similar semantic information.
\textbf{\textit{2) Training instance segmentation models}}: They serve as pseudo ground-truth labels for training a non-agnostic instance segmentor.

\noindent \textbf{Semantic-aware Copy-Paste Augmentation.}
In cluttered natural scenes, previous unsupervised instance segmentation model often fail to distinguish instances from the same semantic class. This results in multiple instances being captured in the same mask.
To distinguish multiple overlapping objects and small objects in existing unsupervised detectors, we employ semantic-aware copy-paste augmentation, which includes several steps:

\textbf{\textit{1)}} We begin by randomly selecting two instances, denoted as $I_1$ and $I_2$, both belonging to the same pseudo-category (or group/cluster).
\textbf{\textit{2)}} One of these instances undergoes a transformation function $\mathcal{T}$, which randomly resizes and shifts the associated pseudo-masks.
\textbf{\textit{3)}} The resized instance is then pasted onto another image, creating synthetic overlapping cases using the following equation:
\begin{align}
    I_3 = I_1 \cdot (1 - \mathcal{T}({M}_c)) + I_2 \cdot \mathcal{T}({M}_c)
\end{align}
where $\cdot$ denotes element-wise multiplication.

\noindent \textbf{Learning Unsupervised Instance Segmentor.} 
Traditionally, unsupervised segmentation community focused primarily on class-agnostic instance segmentation~\cite{wang2022freesolo,wang2022tokencut,wang2023cut}, whose outputs lack class labels. However, by incorporating clustering information obtained from pseudo-labels on ImageNet, as discussed above, our method allows the model to predict not only the location and segmentation of an object but also its pseudo semantic labels. 

As observed by~\cite{wang2023cut}, ``ground-truth'' masks may miss instances. However, a standard detection loss penalizes predicted regions $r_i$ that do not overlap with the ``ground-truth''.
Therefore, following \cite{wang2023cut}, 
we drop the loss for each predicted region $r_i$ that has a maximum overlap of $\tau^{\text{IoU}}$ with any of the `ground-truth' instances:
$\mathcal{L}_{\text{drop}}(r_i) = \mathbbm{1}(\text{IoU}_i^{\text{max}} > \tau^{\text{IoU}})\mathcal{L}_{\text{vanilla}}(r_i)$,
\noindent where $\text{IoU}_i^{\text{max}}$ denotes the maximum IoU with all `ground-truth' for $r_i$ and $\mathcal{L}_{\text{vanilla}}$ is the vanilla loss function of detectors.
$\mathcal{L}_{\text{drop}}$ encourages the exploration of image regions missed in the ``ground-truth''.


\tabComparisonlarge{!t}

\subsection{Unsupervised Universal Image Segmentation}
\label{sec:uni-cutler}



\paragraph{Pseudo Labels for Panoptic Segmentation.}
For each pixel $(i,j)$ in the image, we vector quantize pixels with different semantics or instance membership, generating pseudo semantic labels for panoptic segmentation. 
We assign each pixel a semantic label based on ``stuff'' or ``things'' identity. This results in an instance label $(I(i, j))$ for ``things'' or a semantic label $(S(i, j))$ for ``stuff''. 
The critical challenge in this process is distinguishing between pixels associated with ``things'' (countable, often foreground) and "stuff" (uncountable, usually background)~\cite{adelson2001seeing}.

To resolve this problem, our method unfolds in three steps:
\textit{\textbf{1) Semantic Labeling for ``Things'':}} Utilizing the class-agnostic instance segmentation capabilities of CutLER~\cite{wang2023cut}, we first identify ``things'' within an image, generating class-agnostic instance masks. 
These masks then undergo deep clustering to attribute a semantic label \(I_{C}(i,j)\) to each instance, detailed in~\cref{sec:g-cutler}.
\textit{\textbf{2) Semantic Labeling for ``Stuff'':}} For ``stuff'' pixels, we deploy the unsupervised semantic segmentation model STEGO~\cite{hamilton2022unsupervised}, which distills DINO’s unsupervised features into discrete semantic labels, as outlined in~\ref{sec:preliminaries}. This step assigns a ``stuff'' semantic label to all pixels, including those of ``Things'' identified earlier.
\textit{\textbf{3) Integrating Labels for ``Things'' and ``Stuff''.}} We determine a pixel's classification as ``things'' or ``stuff'' using the following logic:
\begin{align}
    I(i, j) &= \begin{cases}
        I_C(i, j), & \text{if } I_C(i, j)\!\neq\!0 \\
        S_S(i, j), & \text{if } I_C(i, j)\!=\!0 \text{ \& } S_S(i, j)\!\neq\!0\\
        0, & \text{otherwise}
    \end{cases}
\end{align}
This process merges the semantic labels, assigning priority to ``things'' labels over ``stuff'' where applicable.
We then train a universal segmentation model on these pseudo-labels for instance, semantic and panoptic segmentation tasks.

\noindent \textbf{Learning Unsupervised Universal Image Segmentor.} After we obtain the pseudo labels for panoptic segmentation, following~\cite{kirillov2019panopticfpn}, we construct an unsupervised universal image segmentation model, that has two branches: instance segmentation branch and semantic segmentation branch, to address corresponding segmentation tasks. The model is trained jointly for both branches, employing the following loss function: $\mathcal{L} = \lambda_i(\mathcal{L}_{c} + \mathcal{L}_{b} +\mathcal{L}_{m}) + \lambda_s \mathcal{L}_{s}$,
where $\mathcal{L}_{c}$ represents the classification loss, $\mathcal{L}_{b}$ is the detection loss, $\mathcal{L}_{m}$ is the segmentation loss, and $\mathcal{L}_s$ signifies the semantic loss. The $\mathcal{L}_s$ is computed as a per-pixel cross-entropy loss between the predicted and ground-truth labels. The hyperparameters $\lambda_i$ and $\lambda_s$ balance these two parts.

\def\tabComparison#1{
\begin{table}[#1]
    \tablestyle{1.5pt}{1.0}
    \small
    \begin{center}
    \begin{tabular}{l|cc|cc|cc}
    Task $\rightarrow$            & \multicolumn{2}{c|}{Agnostic Ins Seg} & \multicolumn{2}{c|}{Ins Seg} & \multicolumn{2}{c}{Sem Seg} \\ 
    Metric $\rightarrow$          & AP$^\text{box} $                     & AP$_{50}^\text{box}$            &AP$_{50}^\text{box}$        & AR$_{100}^\text{box}$         & PixelAcc            & mIoU           \\ \shline
    FreeSOLO~\cite{wang2022freesolo}        & 9.6                     & 4.2                     & -              & -               & -               & -              \\
    TokenCut~\cite{wang2022tokencut}        & 5.8                     & 3.0                     & -              & -               & -               & -              \\
    CutLER~\cite{cutler}  & 21.9                    & 12.3                    & -              & -               & -               & -              \\
    DINO~\cite{caron2021emerging}            & -                       & -                       & -              & -               & 30.5            & 9.6            \\
    PiCIE + H   \citep{Cho2021PiCIEUS}         & -                       & -                       & -              & -               & 48.1            & 13.8           \\
    STEGO \citep{hamilton2022unsupervised}           & -                       & -                       & -              & -               & 56.9            & 28.2           \\ 
    \hline
    \rowcolor{gray!10}
    \Ours            & \textbf{22.8}                   & \textbf{13.0}                    & \textbf{11.8}           & \textbf{21.5}            & \textbf{63.9}            & \textbf{30.2}           \\
    \rowcolor{gray!10}
    \textit{vs.prev.SOTA}    & \textbf{\color{ForestGreen}{+0.9}}                    & \textbf{\color{ForestGreen}{+0.7}}                     & -              & -               & \textbf{\color{ForestGreen}{+7.0}}            & \textbf{\color{ForestGreen}{+2.0}}           \\ 
    \end{tabular}\vspace{-12pt}
    \caption{The \textbf{comparisons of the proposed methods with the SOTA unsupervised methods.} We report the comparison between the proposed method with the SOTA methods separately for each sub-task on COCO \texttt{val2017}. }
    \label{tab:comp}
    \end{center}
\end{table}
}

\def\FigClusteringVis#1{
\begin{figure}[#1]
  \centering
  \vspace{-10pt}
  \includegraphics[width=0.88\linewidth]{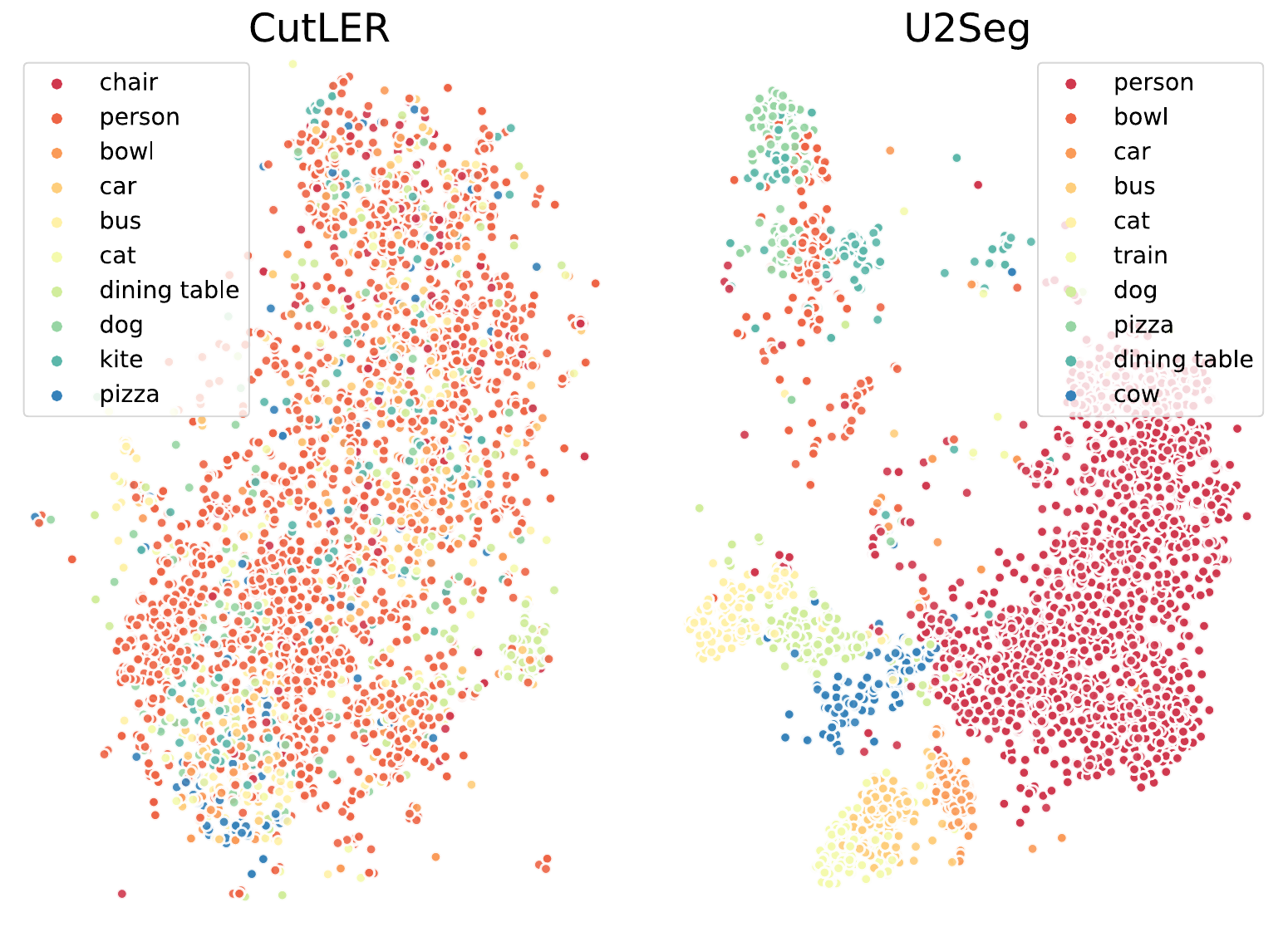}\vspace{-14pt}
  \caption{
  \Ours learns features that are more discriminative than those learned by CutLER.
  The \textbf{t-SNE}~\cite{van2008visualizing} visualization of the features from the model's FC layer. We color-code each dot based on its ground-truth category.}
  \label{fig:cluster_vis}
\end{figure}
}

\def\personHungarian#1{
\begin{figure*}[#1]
  \centering
  \includegraphics[width=1\linewidth]{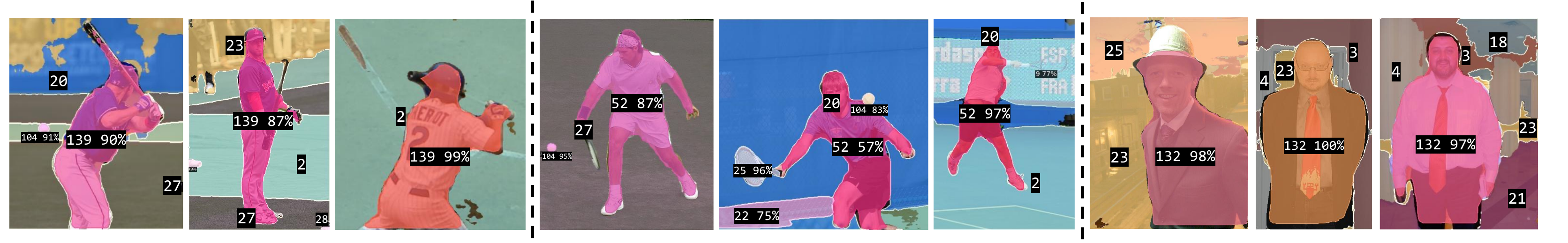}\vspace{-10pt}
  \caption{\textbf{Universal image segmentation visualization} in COCO \texttt{val2017}. We show the results with predicted cluster ID directly from the model, with athletes playing hockey (left columns) as ``139'', playing badminton (middle columns) as ``52'' and the gentlemen (right columns) as ``132''.  After Hungarian matching, these IDs are automatically matched to the category ``person'' for subsequent quantitative evaluation.}
  \label{fig: person_multiple}
\end{figure*}
}

\def\tabInstanceResultsVOC#1{
\begin{table}[!t]
    \tablestyle{10.pt}{1.0}
    \small
    \vspace{-10pt}
    \begin{center}
    \begin{tabular}{lcccc}
    Methods & AP$^\text{box}$ & AP$^\text{box}_{50}$ & AP$^\text{box}_{75}$ & AR$^\text{box}_{100}$ \\
    \shline
    CutLER+   & 17.1     & 26.8       & 18.1       &27.2       \\
    \hline
    \rowcolor{gray!10}
    \Ours       & 19.0 & 31.0  & 19.5   & 48.1  \\ 
    \rowcolor{gray!10}
    $\Delta$      & \textbf{\color{ForestGreen}{+1.9}}      & \textbf{\color{ForestGreen}{+4.2}}       &\textbf{\color{ForestGreen}{+1.4}}         & \textbf{\color{ForestGreen}{+20.9}}       \\
    \end{tabular}\vspace{-16pt}
    \caption{The results for \textbf{zero-shot unsupervised object detection  on PASCAL VOC \texttt{val2012}}. The model is trained on ImageNet with a cluster number of 800. We compare it with CutLER+, a combination of CutLER and offline clustering.}
    \label{tab: ins_seg_voc}
    \end{center}
    \end{table}
}

\def\tabInstanceResultsUVO#1{
\begin{table}[!t]
    \tablestyle{4pt}{1.0}
    \small
    \vspace{-2pt}
    \begin{center}
    \begin{tabular}{lcccccc}
    Metric   & AP$^\text{box}$ & AP$^\text{box}_{50}$ & AR$^\text{box}_{100}$ & AP$^\text{mask}$ & AP$^\text{mask}_{50}$ & AR$^\text{mask}_{100}$ \\ \shline
    CutLER+ & 6.3 & 10.6 & 11.8 & 6.0 & 9.0 & 10.4      \\
    \hline
    \rowcolor{gray!10} 
    \Ours     & 6.8 & 10.8 & 25.0 & 6.2 & 9.5 & 21.0      \\ 
    \rowcolor{gray!10}
    $\Delta$    & \textbf{\color{ForestGreen}{+0.5}}&\textbf{\color{ForestGreen}{+0.2}} &\textbf{\color{ForestGreen}{+13.2}} &\textbf{\color{ForestGreen}{+0.2}} &\textbf{\color{ForestGreen}{+0.5}} &\textbf{\color{ForestGreen}{+10.6}} \\ 
    \end{tabular}\vspace{-16pt}
    \caption{The results for \textbf{zero-shot unsupervised object detection and instance segmentation on UVO \texttt{val}}. The model is trained on ImageNet with a cluster number of 800. We compare with CutLER+, a combination of CutLER and offline clustering.}
    \label{tab: ins_seg_uvo}
    \end{center}
\end{table}
}

\def\tabInstanceResults#1{
\begin{table}[!t]
    \tablestyle{0.9pt}{1.0}
    \vspace{-8pt}
    \small
    \begin{center}
    \begin{tabular}{lcccccccc}
    Metric                  & AP$^\text{box}$           & AP$^\text{box}_{50}$        & AP$^\text{box}_{75}$       & AR$^\text{box}_{100}$         & AP$^\text{mask}$           & AP$^\text{mask}_{50}$        & AP$^\text{mask}_{75}$       & AR$^\text{mask}_{100}$        \\ 
    \shline
    CutLER+                & 5.9          & 9.0          & 6.1          & 10.3          & 5.3          & 8.6          & 5.5          & 9.3          \\
    \hline
    \rowcolor{gray!10}
    \Ours                    & 7.3          & 11.8         & 7.5          & 21.5          & 6.4          & 11.2         & 6.4          & 18.5         \\ 
    \rowcolor{gray!10}
    $\Delta$ & \textbf{\color{ForestGreen}+1.4} & \textbf{\color{ForestGreen}+2.8} & \textbf{\color{ForestGreen}+1.4} & \textbf{\color{ForestGreen}+11.2} & \textbf{\color{ForestGreen}+1.1} & \textbf{\color{ForestGreen}+2.6} & \textbf{\color{ForestGreen}+0.9} & \textbf{\color{ForestGreen}+9.2} \\
    \end{tabular}\vspace{-16pt}
    \caption{The results for \textbf{zero-shot unsupervised object detection and instance segmentation} on COCO \texttt{val2017}. The model is trained on ImageNet with a cluster number of 800. We compare it with CutLER+, a combination of CutLER and offline clustering.}
    \label{tab: ins_seg}
    \end{center}
\end{table}
}

\def\tabPanopticcityscapes#1{
\begin{table}[#1]
    \tablestyle{8.pt}{1.0}
    \small
    \begin{center}
    \begin{tabular}{llccc}
    Methods        & Pretrain             & PQ & SQ & RQ \\ \shline
    \multicolumn{5}{c}{\textbf{\textit{zero-shot methods}}}                                    \\
    \rowcolor{gray!10} \Ours           & IN            & 15.7   & 46.6   & 19.8   \\ \hline
    \multicolumn{5}{c}{\textbf{\textit{non zero-shot methods}}}                                \\
    CutLER+STEGO       & COCO                 & 12.4       & 36.1       &   15.2     \\
    \hline
    \rowcolor{gray!10}
    \Ours           & COCO                 & 15.4   & 51.5   & 19.0   \\
    \rowcolor{gray!10}
    \Ours           & COCO+IN    & 17.6   & 52.7   & 21.7   \\ 
    \rowcolor{gray!10}
    $\Delta$ & \multicolumn{1}{l}{} & \textbf{\color{ForestGreen}+5.2}  &\textbf{\color{ForestGreen}+16.6} &  \textbf{\color{ForestGreen}+6.5}
    \\
    \end{tabular}\vspace{-16pt}
    \caption{\textbf{Unsupervised Panoptic image segmentation} on Cityscapes \texttt{val}. We show PQ, SQ and RQ on zero-shot and non-zero shot settings with the cluster number of 800. We compare with CutLER+STEGO, a combination of CutLER+ and STEGO.}
    \label{tab: uni_seg_city}
    \end{center}
\end{table}
}

\def\tabPanopticResults#1{
\begin{table}[#1]
    \tablestyle{8pt}{1.0}
    \small
    \vspace{-6pt}
    \begin{center}
    \begin{tabular}{llccc}
    Methods        & Pretrain             & PQ & SQ & RQ \\ \shline
    \multicolumn{5}{c}{\textbf{\textit{zero-shot methods}}}                                    \\
    \rowcolor{gray!10}
    \Ours          & IN      & 11.1                  & 60.1                   & 13.7                  \\
    \hline
    \multicolumn{5}{c}{\textbf{\textit{non zero-shot methods}}}                                \\
    CutLER+STEGO   & COCO          & 12.4                   & 64.9                   & 15.5                  \\
     \hline
     \rowcolor{gray!10}
    \Ours          & COCO          & 15.3                  & 66.5                   & 19.1                  \\
    \rowcolor{gray!10}
    \Ours          & COCO+IN & 16.1                 & 71.1                   & 19.9                  \\ 
    \rowcolor{gray!10}
    $\Delta$ &               & \textbf{\color{ForestGreen}+3.7} &  \textbf{\color{ForestGreen}+6.2}  & \textbf{\color{ForestGreen}+4.4}  \\
    \end{tabular}\vspace{-16pt}
    \caption{\textbf{Unsupervised Panoptic image segmentation on COCO \texttt{val2017}.} 
    We show PQ, SQ and RQ on zero-shot and non-zero shot settings. 
    We use CutLER+STEGO, a combination of CutLER+ and STEGO, as a strong baseline.}
    \label{tab: uni_seg}
    \end{center}
\end{table}
}

\def\tabFineTuneMRCNN#1{
\begin{figure}[#1]
  \centering
  \includegraphics[width=1.0\linewidth]{figures/SSL_Cascade.png}\vspace{-6pt}
  \caption{\textbf{Finetuning \CA for low-shot supervised detection and instance segmentation.}
  We fine-tune a Cascade Mask R-CNN model initialized with \CA or Cutler or MoCo-v2 on varying amounts of labeled data on the COCO dataset.
  We use the same schedule as the self-supervised pretrained MoCo-v2 counterpart and report the detection and instance segmentation performance.
  \CA consistently outperforms the Cutler baseline: in the low-shot setting.
  \CA also outperforms Moco-v2~\cite{chen2020improved}, FreeSOLO~\cite{wang2022freesolo} and DETReg~\cite{bar2022detreg} on this benchmark despite using an older detection architecture. Results with Mask R-CNN are in the appendix.
  }
  \label{fig:fine-tune-coco}
\end{figure}
}

\def\tabTsne#1{
\begin{figure}[#1]
  \centering
  \includegraphics[width=1.0\linewidth]{figures/tsne.jpg}\vspace{-10pt}
  \caption{\textbf{Finetuning \CA for low-shot and fully supervised detection and instance segmentation.}
  We fine-tune a Cascade Mask R-CNN model initialized with \CA or Cutler or MoCo-v2 on varying amounts of labeled data on the COCO dataset.
  }
  \label{fig:tsne}
\end{figure}
}

\def\tabExtraVis#1{
\begin{figure}[#1]
  \centering
  \vspace{-4pt}
  \includegraphics[width=0.95\linewidth]{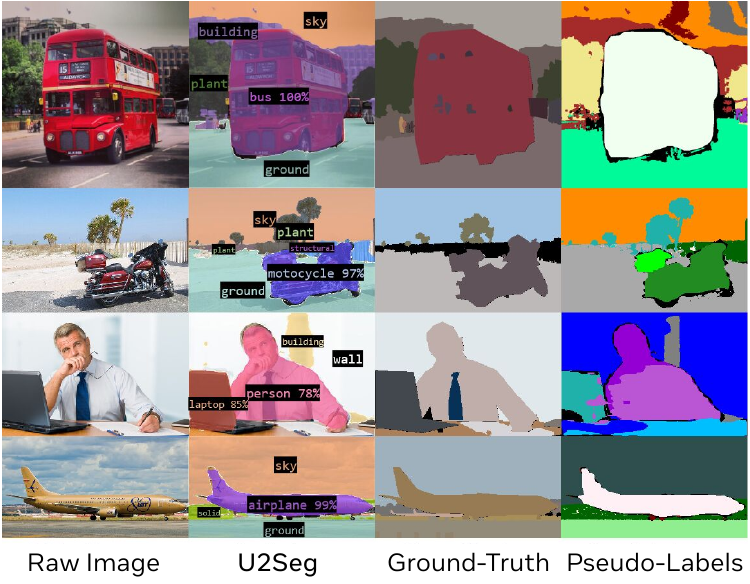}\vspace{-8pt}
  \caption{Visualizations of \Ours's \textbf{unsupervised Panoptic segmentation} results on COCO \texttt{val2017} (\textbf{after Hungarian matching}). 
  The pseudo label is the naive combination of previous state-of-the-art instance segmentation, \ie CutLER~\cite{wang2023cut}, and semantic segmentation, \ie, STEGO~\cite{hamilton2022unsupervised}, results.}
  \label{fig:extra-visual}
\end{figure}
}

\def\tabCityscapesVis#1{
\begin{figure}[#1]
  \centering
  \includegraphics[width=0.95\linewidth]{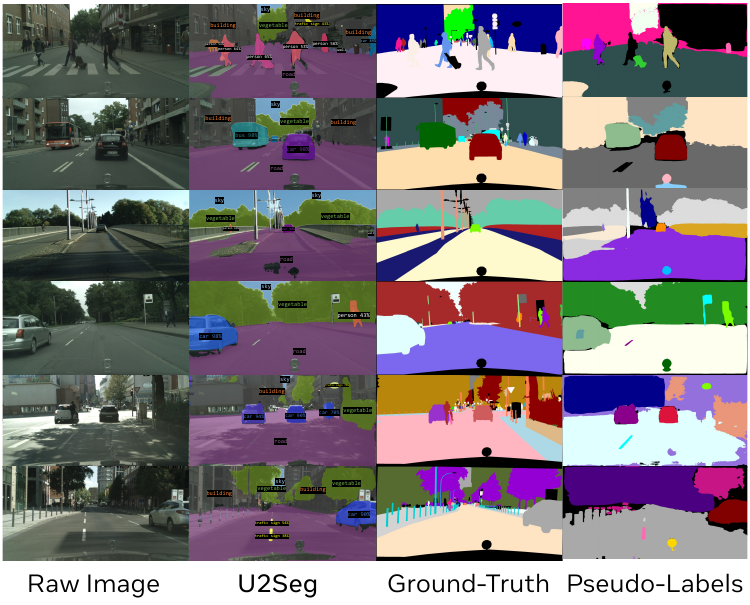}\vspace{-8pt}
  \caption{Qualitative results of \Ours's \textbf{Panoptic image segmentation} results on Cityscapes \texttt{val} (\textbf{after Hungarian matching}).} 
  \label{fig:uni_cityscapes}
\end{figure}
}

\def\figefftwo#1{
\begin{figure*}[#1]
  \centering
  \includegraphics[width=0.95\linewidth]{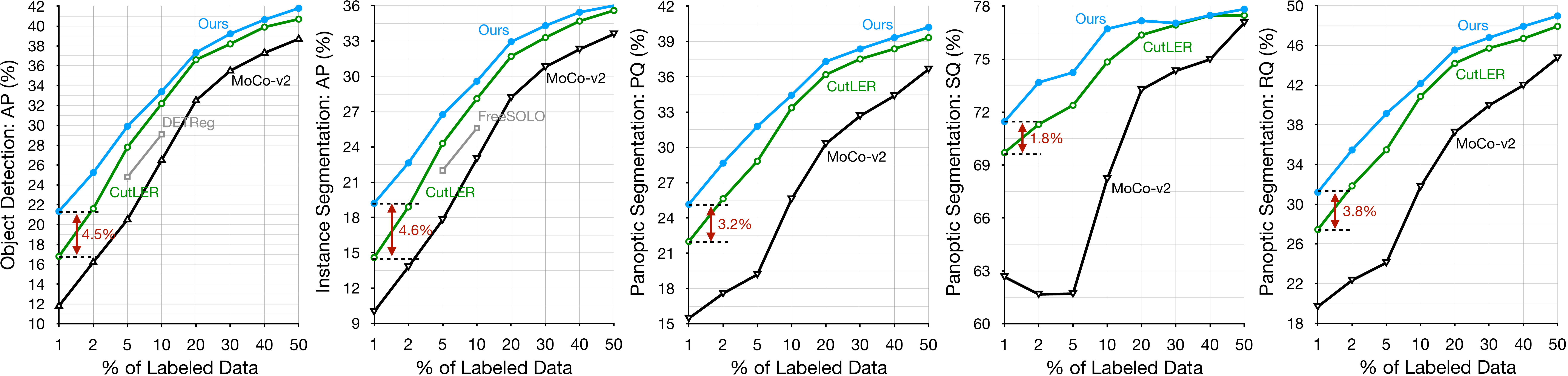}\vspace{-6pt}
  \caption{We evaluate the \textbf{label-efficient learning} performance on 3 different tasks: object detection (the left), instance segmentation (the second left) and panoptic image segmentation (the last three).}
  \label{fig:eff}
\end{figure*}
}

\section{Experiments and Results}
\label{experiments}
\subsection{Experimental Setup} 

\noindent \textbf{Training Data.} Our model is trained on 1.3M unlabeled images from ImageNet~\cite{ImageNet} and is evaluated directly across various benchmarks, unless otherwise noted. For unsupervised semantic segmentation comparisons with STEGO~\cite{hamilton2022unsupervised}, we additionally fine-tune our model using MSCOCO's unlabeled images, following STEGO~\cite{hamilton2022unsupervised}. 

\noindent \textbf{Test Data.} For unsupervised instance segmentation, we test our model on COCO \texttt{val2017}, PASCAL VOC \texttt{val2012}~\cite{everingham2010pascal} and UVO \texttt{val}~\cite{uvo}.
For unsupervised panoptic segmentation, we evaluate our model on COCO \texttt{val2017} and Cityscapes \texttt{val}~\cite{cityscape}.

\noindent \textbf{Evaluation Metrics.} We use AP, AP$_{50}$, AP$_{75}$ and AR$_{100}$ to evaluate the unsupervised instance segmentation; PixelAcc and mIoU for unsupervised semantic segmentation; PQ, RQ, SQ for unsupervised universal image segmentation. After predicting the instance with its semantic labels, we use Hungarian matching to map the semantic labels to class names in the real dataset (details in~\ref{supp: Hungarain}). 
It evaluates the consistency of the predicted semantic segments with the ground truth labels, remaining unaffected by any permutations in the predicted class labels.

\personHungarian{!t}

\noindent \textbf{Implementation Details.} 
Following~\cite{kirillov2019panoptic}, we employ Panoptic Cascade Mask R-CNN \citep{cai2018cascade,kirillov2019panoptic} with a ResNet50 backbone \citep{he2016deep}. 
Following CutLER's training recipe~\cite{wang2023cut}, our model, initialized with DINO pre-trained weights, is trained on unlabeled ImageNet for two epochs. 
It starts with an initial learning rate of 0.01, which then decreases after the first epoch to $5\times10^{-5}$, with a batch size of 16 for all models.
For unsupervised panoptic segmentation, we maintain the same training schedule as unsupervised instance segmentation for zero-shot evaluation. In non-zero-shot scenarios, the models undergo training on a combination of unlabeled COCO and ImageNet datasets, beginning with a learning rate of 0.01 over 90k steps.



\subsection{Unsupervised Universal Image Segmentation}

\label{sec:exp-unuspervised-comparison}
To the best of our knowledge, \Ours represents the first effort in addressing unsupervised \textit{semantic-aware} instance, semantic and panoptic segmentation, all unified under a single framework. 
Due to the absence of benchmarks for unsupervised semantic-aware instance segmentation and panoptic segmentation, we establish comprehensive benchmarks and baselines for both tasks. 


In \cref{tab:comp_full}, we demonstrate that \Ours, utilizing a unified framework, significantly outperforms all previous approaches across various benchmarks and datasets. 
For \textit{\textbf{class-agnostic unsupervised instance segmentation}}, our method achieves an increase of \textbf{+0.9} in AP$^\text{box}$ compared to CutLER~\cite{wang2023cut}. This improvement is largely attributed to our novel semantic-aware copy-paste augmentation, as detailed in \cref{sec:g-cutler}. 
For \textit{\textbf{unsupervised semantic-aware instance segmentation}}, we benchmark against the advanced baseline CutLER+, derived from CutLER, and record a substantial gain of over 11.2\% in AR. A more comprehensive analysis of these results is provided in \cref{sec:exp-unuspervised-ins-seg}.
For \textit{\textbf{unsupervised semantic segmentation}}, our approach surpasses the state-of-the-art STEGO with impressive margins of \textbf{+7.0} in \textbf{PixelAcc} and \textbf{+2.0} in \textbf{mIoU}.
Lastly, for \textit{\textbf{unsupervised panoptic segmentation}}, we compare against the strong baseline of CutLER+STEGO, a hybrid of CutLER+ and STEGO, and observe performance gains of over 6.2\% in SQ on MSCOCO and a notable 16.6\% improvement in SQ on Cityscapes. Further comparisons and discussions on this task are elaborated in \cref{sec:exp-unuspervised-uni-seg}.

\tabInstanceResults{t}

\subsection{Unsupervised Instance Segmentation}
\label{sec:exp-unuspervised-ins-seg}

\tabInstanceResultsVOC{!t}
\tabInstanceResultsUVO{!t}

\tabExtraVis{!bth}

We performed extensive experiments for zero-shot unsupervised instance segmentation. 
Given that prior methods~\cite{wang2023cut,wang2022freesolo,wang2022tokencut,vo2021large} are limited to class-agnostic instance segmentation, we developed CutLER+, a strong baseline for unsupervised semantic-aware instance segmentation, building upon the current state-of-the-art CutLER~\cite{wang2023cut}.
CutLER+ operates in two steps: it first uses the pre-trained CutLER to generate class-agnostic instance masks, and subsequently assigns semantic labels to all instance masks through offline clustering.


~\cref{tab: ins_seg} demonstrates that \Ours markedly improves performance in both unsupervised object detection and instance segmentation on MSCOCO, delivering a \textbf{+2.8} boost in AP$_{50}^\text{box}$ and a \textbf{+2.6} rise in AP$_{50}^\text{mask}$ over CutLER+. 
Additionally, our method sees a substantial increase of approximately \textbf{+10.0} in AR$_{100}$. 
Results on PASCAL VOC~\texttt{val2012} and UVO~\texttt{val} are detailed in ~\cref{tab: ins_seg_voc} and ~\cref{tab: ins_seg_uvo}, respectively. Notably, we achieve gains exceeding \textbf{+20\%} in AR for PASCAL VOC and \textbf{+10\%} for UVO.

\subsection{Unsupervised Panoptic Segmentation}
\label{sec:exp-unuspervised-uni-seg}
\tabPanopticcityscapes{!t}
\tabPanopticResults{!t}
\tabCityscapesVis{!t}

\figefftwo{!t}
\FigClusteringVis{!t}

For unsupervised panoptic/universal image segmentation, our experiments span two scenarios. 
In the zero-shot setting, the model is trained exclusively on unlabeled ImageNet images. 
For non-zero-shot (in-domain) scenarios, we train on unlabeled COCO images or a mix of COCO and ImageNet.
With no existing benchmarks for unsupervised panoptic segmentation, we establish a new baseline by integrating the state-of-the-art unsupervised semantic segmentation from STEGO~\cite{hamilton2022unsupervised} with semantic-aware instance segmentation from CutLER+ (discussed in \cref{sec:exp-unuspervised-ins-seg}), which are then merged to create panoptic/universal segmentation outcomes, referred to as CutLER+STEGO.




~\cref{tab: uni_seg} presents the PQ, SQ, and RQ scores of \Ours on COCO \texttt{val2017}. \Ours surpasses the strong baseline CutLER+STEGO with a \textbf{+3.5} improvement in PQ and an increase of over \textbf{+4.0} in RQ. Qualitative results of \Ours's performance is provided in~\cref{fig: person_multiple}, with the predicted semantic labels visualized. 
The qualitative results suggest that an over-clustering strategy in pseudo-label generation, \eg setting the number of clusters to 300 or 800, leads to highly granular model predictions. 
For instance, as in~\cref{fig: person_multiple}, the model distinctly categorizes hockey players as ``139'', badminton players as ``52'', and gentlemen in suits as ``132'', showcasing its potent discriminative capabilities.

To quantitatively measure the quality of segmentation masks and their corresponding semantic labels, we use Hungarian matching (detailed in~\cref{supp: Hungarain}) to align semantic labels with the category names from the test dataset; for instance, all three sub-clusters depicted in ~\cref{fig: person_multiple} are assigned to the "person" category. The qualitative outcomes post-Hungarian matching are shown in \cref{fig:extra-visual}, where our model demonstrates superior panoptic segmentation mask quality. For instance, while the baseline tends to segment parts separately (as seen with the man's head and torso being treated as separate entities in the third row), our model correctly identifies them as parts of a single object. This level of recognition is also evident with the ``trunks of the motorcycle'' example in the second row. For additional results, please see~\cref{sec: supp_universal}.
We also present results of the more challenging dataset Cityscapes in~\cref{tab: uni_seg_city} and~\cref{fig:uni_cityscapes}.

\subsection{Efficient Learning}
\label{sec:exp-unuspervised-eff}

Specifically, for object detection and instance segmentation, we employ our unsupervised instance segmentation model, with cluster count set to 300, to initialize the model weights. 
We adopt the recipe from \cite{cutler, densecl} for model fine-tuning across various annotation splits. 
For label-efficient panoptic segmentation, we fine-tune the model initialized with our zero-shot unsupervised framework on the same data splits.

The results are depicted in \cref{fig:eff}, where our model's instance segmentation performance is benchmarked against MoCo-V2, DETReg, and CutLER. Our model consistently surpasses the state-of-the-art with consistent gains in both AP$^\text{box}$ and AP$^\text{mask}$. 
In scenarios with panoptic image segmentation as the downstream task, we contrast our results with MoCo-V2 and CutLER in terms of PQ, SQ, and RQ metrics. 
The results illustrate a remarkable improvement, effectively doubling the performance boost from MoCo-V2 to CutLER, especially in few-shot contexts with limited annotations (1\% or 2\% labeled samples). 
This highlights the practical value of our approach in real-world unsupervised learning applications, where annotations are often scarce.



We attribute the performance gains primarily to the discriminative features our model learns, as in~\cref{fig:cluster_vis}, obtaining effective model initialization for few-shot learning. 

\def\clusternum#1{
\begin{table}[#1]
    \tablestyle{5pt}{0.8}
    \small
    \begin{center}
    \begin{tabular}{l|cc|cc|cc}
    \multirow{2}{*}{\# Cluster} & \multicolumn{2}{c|}{COCO} & \multicolumn{2}{c|}{UVO} & \multicolumn{2}{c}{VOC} \\ 
                      & AP$_{50}^\text{box}$          & AR$_{100}^\text{box}$         & AP$_{50}^\text{box}$         & AR$_{100}^\text{box}$         & AP$_{50}^\text{box}$         & AR$_{100}^\text{box}$         \\ \midrule[1.2pt]
    300               & 9.3         & 20.1       & 9.8        & 22.6       & 29.6       & 45.7       \\
    800               & 11.8        & 21.5       & 10.8            &   25.0         & 31.0         & 48.0         \\
    2911              & 13.3        & 22.1       &15.1            &   25.8         & 31.6       & 48.3       \\
    \end{tabular}\vspace{-18pt}
    \caption{Over-clustering can improve the model performance. We show results on different datasets for the unsupervised object detection using \textbf{different cluster numbers}. }
    \label{tab: abl_cluster}
    \end{center}
\end{table}
}

\def\hungrian#1{
\begin{table}[#1]
    \tablestyle{1.2pt}{0.8}
    \small
    \vspace{-8pt}
    \centering
    \begin{minipage}{.45\linewidth}
    \centering
    \subfloat[\textbf{Conf's effect on accuracy.}]{%
    \begin{tabular}{c|cccc}
    conf & \# matched & AP$_{50}^\text{box}$ & AR$_{100}^\text{box}$ \\ 
    \midrule[1.2pt]
    0.9  & 109 & 10.9 & 13.1   \\
    0.7  & 225 & 11.6 & 18.0   \\
    0.6  & 282 & 11.8 & 19.7   \\
    \rowcolor{gray!10}
    \textbf{0.4}  & \textbf{389} & \textbf{11.8} & \textbf{21.5}   \\
    0.2  & 513 & 11.3 & 21.8   \\
    0.0  & 718 & 8.6  & 18.4   \\
    \end{tabular}
    } 
    \end{minipage}
    \hfill
    \begin{minipage}{.45\linewidth}
    \centering
    \subfloat[\textbf{IoU's effect on accuracy.}]{%
    \begin{tabular}{c|cccc}
    IoU & \# matched & AP$_{50}^\text{box}$ & AR$_{100}^\text{box}$ \\ 
    \midrule[1.2pt]
    0.9 & 295 & 10.8 & 19.7   \\
    0.8 & 348 & 11.4 & 20.7   \\
    0.4 & 414 & 11.5 & 21.6   \\
    0.2 & 450 & 11.5 & 21.1   \\
    0.0 & 494 & 9.2  & 17.7   \\
    \rowcolor{gray!10}
    \textbf{0.6} & \textbf{389} & \textbf{11.8} & \textbf{21.5}   \\
    \end{tabular}
    } 
    \end{minipage}\vspace{-8pt}
    \caption{
    \textbf{Impact of Confidence and IoU} on Hungarian Matching Performance: The left table illustrates the outcomes at a fixed IoU of 0.6 while varying the confidence scores. 
    Conversely, the right table displays the results with a constant confidence of 0.4, altering the IoU values. The cluster number is 800. 
    }
    \label{tab: Hungarian}
    \end{table}
}

\subsection{Ablation Studies}
\label{sec:ablation}
In this section, we conduct ablation study on \Ours. 

\clusternum{!t}

\noindent \textbf{Numbers of clusters.} 
The choice of cluster quantity significantly affects the model's representation granularity. Our ablation study on various cluster counts, as detailed in~\cref{tab: abl_cluster}, reveals their impact on model performance. 
Over-clustering generally leads to a finer level of detail, prompting the model to learn more discriminative features.


\noindent \textbf{Hungarian matching.} 
As our trained model could predict the instance with corresponding semantic labels, we are able to go further beyond unsupervised class-agnostic instance segmentation. To quantitatively evaluate the performance, Hungarain matching is employed to match the predicted semantic labels to the ground-truth dataset categories. See~\cref{supp: Hungarain} for details of the adopted Hungarian matching used in our evaluation.
As shown in~\cref{tab: Hungarian}, the two parameters conf threshold and IoU threshold also affect the precision and recall.

\hungrian{!t}

\section{Summary}




We present \textbf{\Ours}, a novel \textbf{U}nsupervised \textbf{U}niversal Image \textbf{Seg}mentation model, adept at performing unsupervised instance, semantic, and panoptic segmentation tasks within a unified framework. 
Evaluated on extensive benchmarks, \Ours consistently outperforms previous
state-of-the-art methods designed for individual tasks. 
Additionally, \Ours achieves the new state-of-the-art for label-efficient panoptic segmentation and instance segmentation.
We anticipate that \Ours, free from the constraints of human annotations, will demonstrate enhanced performance when scaled up with more training data, representing an important direction for our future research.

\section{Acknowledgement}
Trevor Darrell, Dantong Niu and Xudong Wang were funded by DoD including DARPA LwLL and the Berkeley AI Research (BAIR) Commons.


\def\uniresults#1{
\begin{table*}[#1]
\tablestyle{9.pt}{1.0}
\small
\begin{center} 
\begin{tabular}{l|l|c|ccccccccc}
 Datasets  &Pretrain & \# Cluster & PQ      & PQ$^\text{St}$     & PQ$^\text{Th}$     & SQ      & SQ$^\text{Th}$     & SQ$^\text{St}$    & RQ       & RQ$^\text{Th}$      & RQ$^\text{St}$     \\ \shline
 \multirow{6}{*}{COCO} &IN              & 300        & 11.1 & 9.5  & 19.3 & 60.1 & 60.3 & 59.0 & 13.7   & 11.6 & 25.0 \\
       &IN                             & 800        & 11.9 & 10.5 & 19.6 & 65.9 & 67.4 & 58.2 & 14.8   & 12.8 & 25.3 \\
     &COCO                             & 300        & 15.3 & 14.2 & 21.6 & 66.5 & 67.2 & 62.4 & 19.1   & 17.5 & 27.5 \\
     &COCO                             & 800        & 15.5 & 14.6 & 20.5 & 69.7 & 71.1 & 62.6 & 19.1   & 17.8 & 26.1 \\
  &IN+COCO                             & 300        & 15.5 & 14.4 & 21.2 & 67.1 & 67.7 & 64.3 & 19.2   & 17.8 & 26.9 \\
  &IN+COCO                             & 800        & 16.1 & 15.1 & 21.2 & 71.1 & 72.5 & 63.8 & 19.9   & 18.6 & 26.8 \\ \hline 
\multirow{6}{*}{Cityscapes} &IN & 300        & 15.3 & 4.1  & 23.4 & 48.8 & 54.7 & 44.6 & 19.5   & 5.4  & 29.7 \\
       &IN                             & 800        & 15.7 & 4.3  & 24.0 & 46.6 & 47.5 & 45.9 & 19.8 & 5.5  & 30.2 \\
     &COCO                             & 300        & 18.4 & 7.8  & 26.1 & 47.4 & 47.3 & 47.4 & 22.6   & 9.8  & 31.9 \\
     &COCO                             & 800        & 15.4 & 5.8  & 22.3 & 51.5 & 62.9 & 43.2 & 19.0   & 7.5  & 27.4 \\
  &IN+COCO                             & 300        & 16.5 & 6.2  & 24.1 & 44.1 & 45.2 & 43.3 & 20.5   & 7.9  & 29.7 \\
  &IN+COCO                             & 800        & 17.6 & 8.4  & 24.2 & 52.7 & 67.5 & 42.0 & 21.7   & 10.5 & 29.9 \\ 
\end{tabular}\vspace{-1.2em}

\end{center}
\caption{Complete results for \textbf{unsupervised universal image segmentation.} We show results for different models pretrained on various dataset and test on COCO \texttt{val2017}, Cityscapes \texttt{val}, with corresponding cluster numbers.}
\label{supp: tab-universal}
\end{table*}
}

\def\FigClusteringVis#1{
\begin{figure*}[#1]
  \centering
  \includegraphics[width=0.95\linewidth]{figures/clustering.jpg}
  \caption{Visualization of clustering.}
  \label{fig:cluster_vis}
\end{figure*}
}

\def\FigBackground#1{
\begin{figure*}[#1]
  \centering
  \includegraphics[width=0.9\linewidth]{figures/mis-seg-background.pdf}
  \caption{Clusters of the instances provided by Cutler that have more than 10\% of the images that have at least 2 corners in the instance mask.}
  \label{fig:extra-visual}
\end{figure*}
}

\def\FigInstance#1{
\begin{figure*}[#1]
  \centering
  \includegraphics[width=\linewidth]{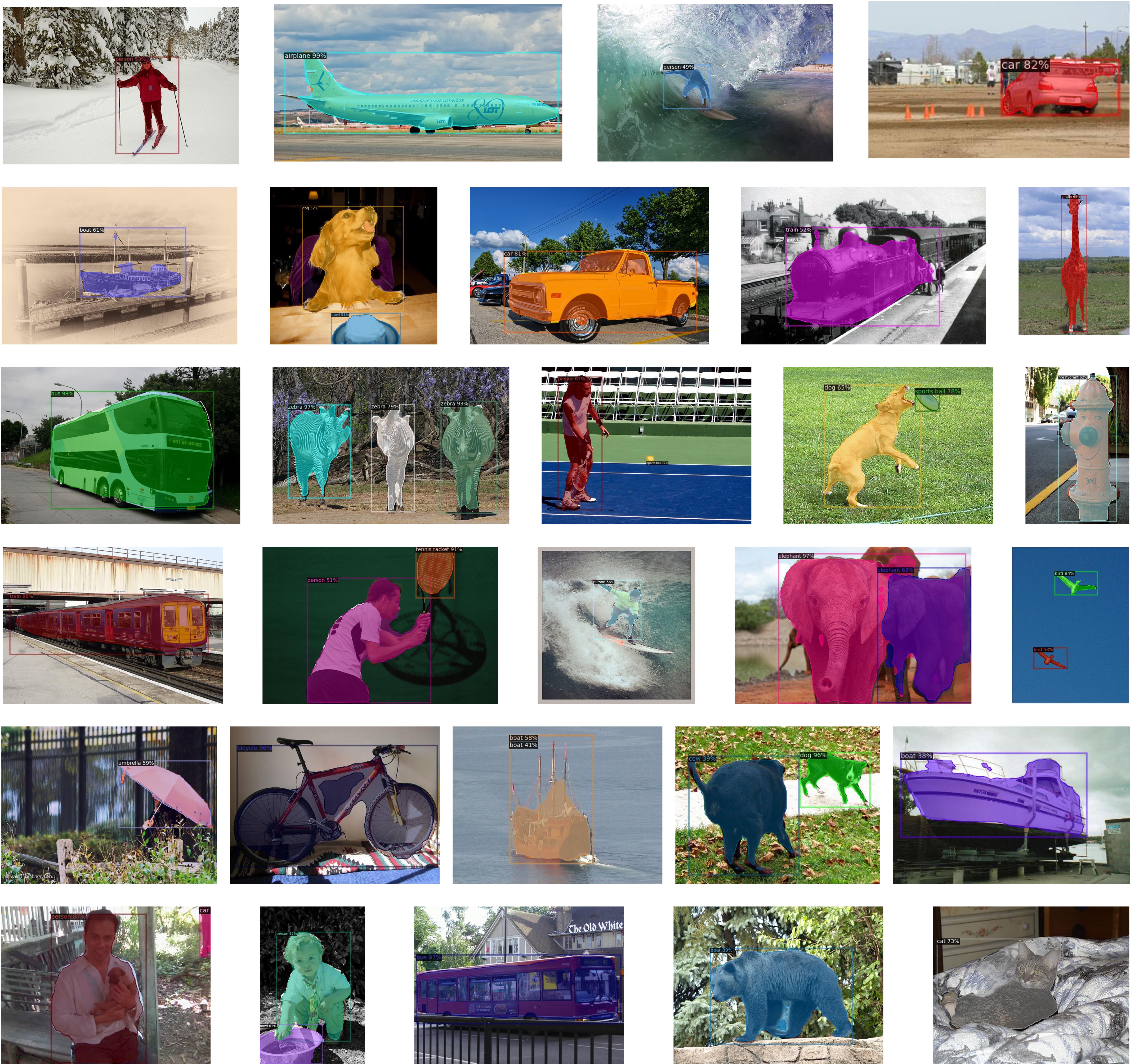}
  \caption{\textbf{Unsupervised object detection and instance segmentation visualization} of COCO \texttt{val2017} and PASCAL VOC \texttt{val2012} (\textbf{after Hungarian matching}).}
  \label{supp: instance_vis}
\end{figure*}
}

\def\FigCityscapeVis#1{
\begin{figure}[#1]
  \centering
  \includegraphics[width=\linewidth]{figures/cityscape.pdf}
  \caption{Unsupervised universal image segmentation visualizations of Cityscape dataset.}
  \label{supp: cityscape_universal_vis}
\end{figure}
}

\def\FigUniversal#1{
\begin{figure*}[#1]
  \centering
  \includegraphics[width=0.73\linewidth]{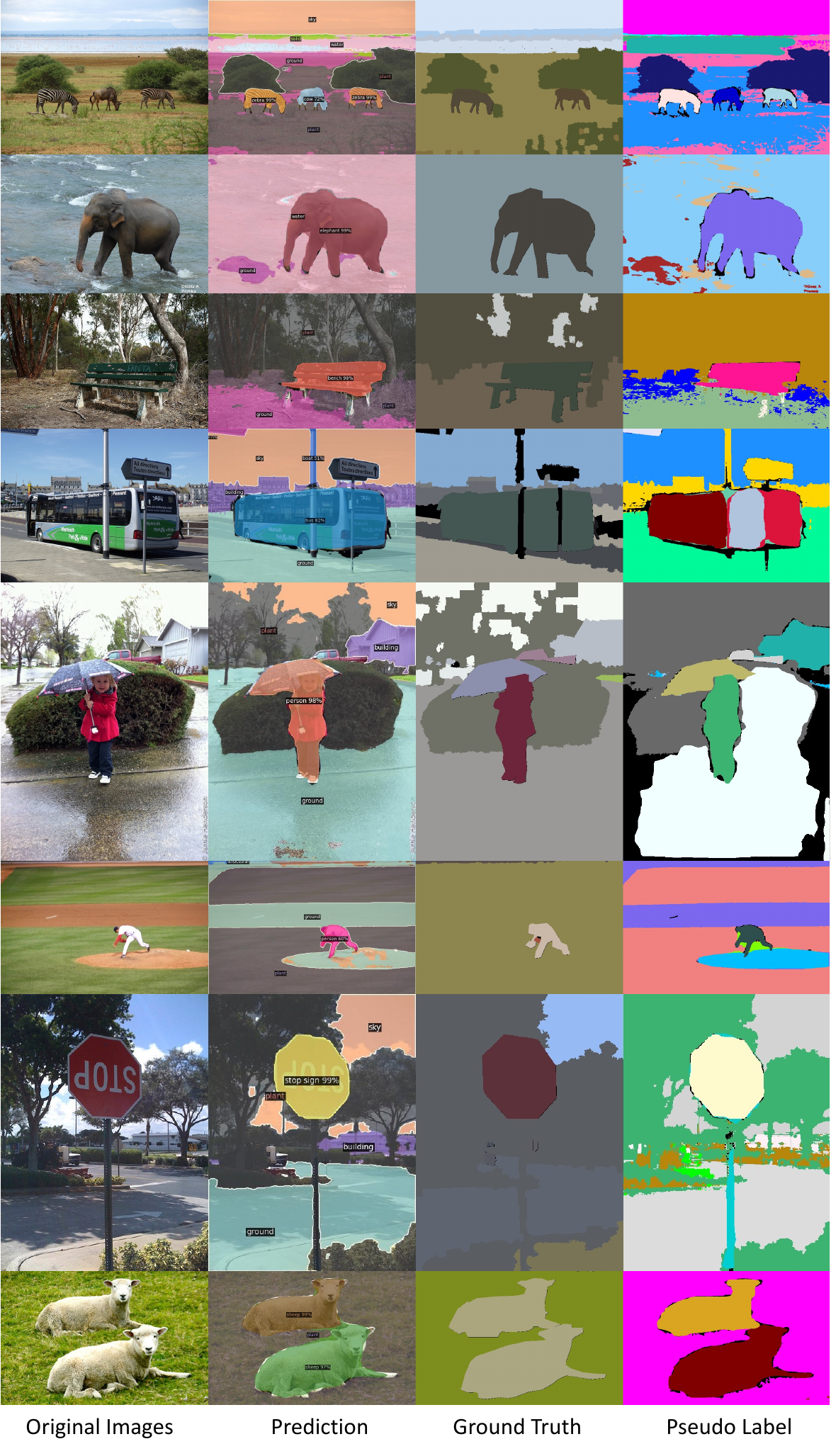}
  \caption{\textbf{Unsupervised universal image segmentation} visualizations of COCO \texttt{val2017} (\textbf{after Hungarian matching}).} 
  \label{supp: universal_vis}
\end{figure*}
}

\def\tabDatasets#1{
  \begin{table*}[#1]
  \tablestyle{15pt}{1.0}
  \small
  \begin{center}
  \begin{tabular}{lccccc}
  Datasets & Domain & Testing Data & \#Images & Instance Segmentation Label \\ \shline
  COCO~\cite{lin2014microsoft} & natural images & \texttt{val2017} split & 5,000 & \cmark \\
  UVO~\cite{wang2021unidentified} & video frames & \texttt{val} split & 21,235 & \cmark \\
  PASCAL VOC~\cite{everingham2010pascal} & natural images & \texttt{trainval07} split & 9,963 & \xmark \\
  Cityscapes~\cite{cityscape} & urban scenes & \texttt{val} split & 500 &  \cmark \\
  \end{tabular}
  \end{center}
  \caption{\textbf{Summary of datasets} used for evaluation.}
  \label{tab:dataset-summary}
  \end{table*}
}

\def\tabBoxPerformance#1{
\begin{table*}[#1]
\tablestyle{7.pt}{1.0}
    \small
\centering
\begin{tabular}{l|ccc|ccccccccc}
Datasets  & \# cluster & IoU & Conf & AP$^\text{box}$ & AP$_{50}^\text{box}$ & AP$_{75}^\text{box}$ & AP$_{S}^\text{box}$ & AP$_{M}^\text{box}$ & AP$_{L}^\text{box}$ & AR$_{1}^\text{box}$ & AR$_{10}^\text{box}$ & AR$_{100}^\text{box}$ \\
\shline
\multirow{3}{*}{UVO} &2911 & 0.6 & 0.1 & 9.7 & 15.1 & 9.3 & 0.6 & 5.2 & 14.4 & 18.0 & 25.3 & 25.8 \\
 &800 & 0.4 & 0.1 & 6.8 & 10.8 & 7.2 & 0.6 & 2.9 & 10.2 & 17.2 & 24.5 & 25.0 \\
 &300 & 0.7 & 0.1 & 6.5 & 9.8 & 6.5 & 0.8 & 2.6 & 9.2 & 16.0 & 22.2 & 22.6 \\ \hline
\multirow{3}{*}{VOC} &2911 & 0.5 & 0.2 & 19.2 & 31.6 & 19.7 & 1.0 & 6.4 & 26.6 & 28.6 & 44.9 & 48.3 \\
 &800 & 0.8 & 0.2 & 19.0 & 31.0 & 19.5 & 0.6 & 4.8 & 26.6 & 28.8 & 45.2 & 48.1 \\
 &300 & 0.8 & 0.4 & 18.4 & 29.6 & 18.8 & 0.3 & 3.8 & 26.0 & 27.1 & 41.0 & 42.8 \\ \hline
\multirow{3}{*}{COCO} &2911 & 0.5 & 0.3 & 8.2 & 13.3 & 8.4 & 1.4 & 7.0 & 18.2 & 14.1 & 21.4 & 22.1 \\
 &800 & 0.6 & 0.4 & 7.3 & 11.8 & 7.5 & 1.2 & 5.8 & 15.8 & 13.3 & 20.8 & 21.5 \\
  &300 & 0.6 & 0.3 & 5.7 & 9.3 & 5.9 & 0.5 & 4.6 & 12.9 & 11.9 & 19.5 & 20.1 \\
\end{tabular}
\caption{\textbf{Complete results for unsupervised object detection.} We show results on UVO \texttt{val}, PASCAL VOC \texttt{val2012} and COCO \texttt{val2017}, with corresponding clustering numbers. The IoU and Conf are the Hungarian matching parameter we use for evaluation.}
\label{tab:box-performance}
\end{table*}
}

\def\tabMaskPerformance#1{
\begin{table*}[#1]
\tablestyle{6.pt}{1.0}
    \small
\centering
\begin{tabular}{l|ccc|ccccccccc}

Datasets & \# cluster & IoU & Conf & AP$^\text{mask}$ & AP$_{50}^\text{mask}$ & AP$_{75}^\text{mask}$ & AP$_{S}^\text{mask}$ & AP$_{M}^\text{mask}$ & AP$_{L}^\text{mask}$ & AR$_{1}^\text{mask}$ & AR$_{10}^\text{mask}$ & AR$_{100}^\text{mask}$ \\
\shline
\multirow{3}{*}{UVO} &2911 & 0.6 & 0.1 & 8.8 & 13.9 & 8.4 & 0.5 & 6.4 & 14.4 & 16.0 & 21.7 & 22.1 \\
 &800& 0.4 & 0.1 & 6.2 & 9.5 & 6.0 & 0.5 & 2.1 & 9.8 & 15.7 & 20.6 & 21.0 \\ 
 &300 & 0.7 & 0.1 & 6.1 & 9.5 & 5.8 & 0.7 & 1.0 & 8.8 & 14.1 & 19.2 & 19.4 \\ \hline
\multirow{3}{*}{COCO} &2911 & 0.5 & 0.3 & 7.3 & 12.4 & 7.4 & 0.8 & 4.9 & 17.9 & 12.8 & 18.7 & 19.2 \\
 &800 & 0.6 & 0.4 & 6.4 & 11.2 & 6.4 & 0.7 & 3.7 & 15.0 & 11.9 & 18.0 & 18.5 \\
 &300 & 0.6 & 0.3 & 4.9 & 8.6 & 5.0 & 0.3 & 2.6 & 11.8 & 10.7 & 16.9 & 17.3 \\
\end{tabular}
\caption{Complete results for \textbf{unsupervised instance segmentation.} We show results on UVO \texttt{val} and COCO \texttt{val2017}, with corresponding clustering numbers. The IoU and Conf is the Hungarian matching parameter we use for evaluation.}
\label{tab:mask-performance}
\end{table*}

}

\def\tabLimitaton#1{
\begin{table}[#1]
    \tablestyle{9.pt}{1.0}
    \small
    \begin{center}
    \begin{tabular}{l|cccc}
    Model   & AP$^\text{box}$ & AP$^\text{box}_{50}$  & AP$^\text{mask}$ & AP$^\text{mask}_{50}$ \\ \shline
    CutLER+  & 5.9  & 9.0    & 5.3  &8.6  \\ \hline 
    \rowcolor{gray!10} Panoptic & 6.1 & 9.8 & 5.8 &9.0 \\
    \rowcolor{gray!10} Instance  & 7.3  & 11.8 & 6.4 &11.2  \\
    \end{tabular} \vspace{-1.2em}
    \caption{\textbf{Limitation of \Ours.} We show the zero-shot unsupervised instance segmentation results on COCO \texttt{val2017}. CutLER+ is evaluated on the combination of CutLER and offline clustering, Panoptic is trained on both \textit{``stuff''} and \textit{``things''} pseudo labels, Instance is trained solely on \textit{``things''} labels. }
    \label{tab:limitation}
    \end{center}
\end{table}
}

\appendix
\setcounter{page}{1}
\renewcommand{\thefigure}{A\arabic{figure}}
\setcounter{figure}{0}
\renewcommand{\thetable}{A\arabic{table}}
\setcounter{table}{0}


\vspace{10pt}

\noindent \textbf{\Large Appendix Materials}

\tabDatasets{!htp}

\section{Datasets used for Evaluation}
We provide information about the datasets used in this work as shown in~\cref{tab:dataset-summary}

\noindent  \textbf{\textit{COCO}.} The COCO dataset, introduced by~\cite{lin2014microsoft}, is used for object detection and instance segmentation. It has 115,000 training images, 5,000 validation images, and a separate batch of 123,000 unannotated images. We test our unsupervised instance segmentation on the COCO \texttt{val2017} set with zero-shot setting. We report results using standard COCO metrics, including average precision and recall for detection and segmentation. Also, for unsupervised universal image segmentation, we test the performance on COCO \texttt{val2017}. We report results using panoptic segmentation COCO metrics.

\noindent  \textbf{\textit{PASCAL VOC}.} The PASCAL VOC dataset~\cite{everingham2010pascal} is a widely-used benchmark for object detection. We test our model using the \texttt{trainval07} split and adopt COCO-style evaluation metrics.

\noindent  \textbf{\textit{UVO}.} The UVO dataset~\cite{wang2021unidentified} is designed for video object detection and instance segmentation. We test our unsupervised instance segmentation on the UVO \texttt{val} split, which includes 256 videos with each one annotated at 30 fps. We remove the extra 5 non-COCO categories which are marked as ``other'' in their official annotations. For evaluation, we employ COCO-style metrics.

\noindent  \textbf{\textit{Cityscapes}.} Cityscapes is a dataset dedicated to semantic urban scene understanding, focusing primarily on semantic segmentation of urban scenes. In our research, we tested our unsupervised universal image segmentation on the Cityscapes \texttt{val} splits, using COCO-stype panoptic evaluation metrics.



\section{Hungarian Matching for Unsupervised Segmentation Evaluation}
\label{supp: Hungarain}

In unsupervised object detection and instance segmentation, category IDs are predicted without referencing any predefined labels. For convenience, we differentiate the predicted category ID of \Ours as ``cluster ID'' while keep the ground truth category ID as ``category ID'' in the following analysis. To evaluate the segmentation performance, particularly concerning category accuracy, an optimal correspondence between the cluster ID and the ground truth category ID is essential. We leverage a multi-to-one Hungarian matching for evaluation of \Ours.

\noindent  \textbf{Hungarain Matching.} Given a set of predicted bounding boxes, masks associated with predicted cluster IDs and the corresponding ground truth, the objective is to find the best match from ``cluster ID'' to ``category ID''. To do this, we first use the predicted confidence score conf as a threshold to filter the predicted instance, removing the ones with low confidence. Then, for each predicted instance with its cluster ID, we calculate the IoU of the predicted bounding box or mask with all ground truth instances, then select the one whose IoU is bigger than the predefined threshold, regarding it as the ground truth category ID for this cluster ID. After we get these cluster ID and ground truth category ID pairs, we form a histogram for each kind of cluster ID based on its overlap with all kinds of ground truth category ID. The ground truth category ID that appears most frequently in this histogram becomes the mapping for this cluster ID. This process may result in multiple predicted cluster IDs being mapped to the same ground truth category ID, leading to a multi-to-one matching scenario.

In our experiment, the confidence score threshold conf to filter the predicted instance and the IoU threshold to match predicted instance with its ground truth instance are both hyperparameters, some ablations can be found in Sec. 4.6.





\noindent  \textbf{Evaluation Implications.} The multi-to-one Hungarian matching method provides a systematic and efficient way to assess the performance of unsupervised segmentation models. By mapping predicted cluster ID to their most likely ground truth counterparts, the method ensures that the evaluation reflects the true categorization capability of the model. This, in turn, allows for a fair and consistent comparison across different unsupervised segmentation techniques.

\tabBoxPerformance{!t}
\tabMaskPerformance{!t}
\uniresults{!t}
\section{Unsupervised Instance Segmentation}

In this section, we provide complete results for the unsupervised instance segmentation of \Ours. The results are presented over various datasets and classes to furnish a comprehensive evaluation of our model's capability.

~\cref{tab:box-performance} and~\cref{tab:mask-performance} display the results for unsupervised object detection and instance segmentation on different datasets. One trend can be observed across the different datasets: as the number of the predicted cluster ID increases (\eg, moving from 300 to 2911), there is a consistent increase for most of the metrics. This trend can be succinctly attributed to the intrinsic properties of the multi-to-one Hungarian matching approach (we also show the parameter IoU and Conf used for Hungarian matching). With an increase of the cluster numbers, the Hungarian matching has a broader set of predictions to associate with a single label. This inherently increases the chances of having at least one correct prediction for the given label, making the matching process more amenable. In essence, larger cluster numbers afford easier matching, thereby boosting the evaluation metrics.

Furthermore, the qualitative results are shown in~\cref{supp: instance_vis}, with the samples selected in COCO \texttt{val2017} and PASCAL VOC \texttt{val2012}. After Hungarian matching, we are able to get the real categories of the predicted instances.

\FigInstance{!t}

\section{Unsupervised Universal Image Segmentation}
\label{sec: supp_universal}
Our model's performance for unsupervised universal image segmentation closely mirrors the trends observed in instance segmentation. Specifically, as the number of the predicted clusters increases, the performance of the panoptic segmentation also improves. Detailed universal segmentation results are shown in~\cref{supp: tab-universal} and~\cref{supp: universal_vis}.



\section{Limitation}

\tabLimitaton{!t}

The primary goal of our research is to develop a comprehensive model capable of excelling in all areas of unsupervised segmentation. As shown in~\cref{tab:limitation}, in terms of the individual sub-task, the universal model exhibits a slight underperformance compared to its counterpart model trained with task-specific annotations. This suggests that \Ours is adaptable to various tasks, yet it requires task-specific training to achieve the best outcomes for a specific sub-task. Looking ahead, we aim to develop a more versatile model that can be trained once to effectively handle multiple tasks.

\FigUniversal{!t}

\clearpage

{
\small
\bibliographystyle{ieeenat_fullname}
\bibliography{main}
}


\end{document}